%% file: 029-main.tex
\documentclass[runningheads]{llncs}
\RequirePackage{silence}  
\usepackage{graphicx}
\usepackage{comment}
\usepackage{amsmath,amssymb}
\usepackage{color}
\usepackage{url}
\usepackage[hidelinks]{hyperref}
\usepackage{cleveref}

\usepackage{xcolor}
\usepackage{multicol,multirow}
\usepackage{overpic}
\usepackage[absolute,overlay]{textpos}
\usepackage{float}
\usepackage{wrapfig}
\usepackage{amsmath}
\usepackage{graphicx}
\usepackage{amssymb}
\usepackage{booktabs}
\usepackage{afterpage}
\usepackage{pifont}% http://ctan.org/pkg/pifont
\usepackage{wrapfig}
\usepackage{makecell}
\usepackage{dsfont}
\usepackage{nccmath}
\usepackage{mathtools}
\usepackage{nicefrac}
\usepackage{listings}
\usepackage{rotating}
\usepackage[accsupp]{axessibility} 

\usepackage{soul}
\usepackage{array}
\usepackage{makecell}
\newcolumntype{C}[1]{>{\centering\arraybackslash}m{#1}}
\usepackage{caption}
\usepackage{diagbox}
\usepackage{ulem}
\usepackage{comment}
\usepackage{orcidlink}

\usepackage[sort]{cite}
\usepackage[dvipsnames]{xcolor}

\newcommand{\target}[0]{\mathbf{S}}
\newcommand{\denoiser}[0]{{D}_{\phi}}

\newcommand{\methodname}[0]{ODin}

\renewcommand{\vec}[1]{\boldsymbol{#1}}

\newcommand{\methodnetwork}[0]

\newcommand\reallywidehat[1]{%
\savestack{\tmpbox}{\stretchto{%
  \scaleto{%
    \scalerel*[\widthof{\ensuremath{#1}}]{\kern-.6pt\bigwedge\kern-.6pt}%
    {\rule[-\textheight/2]{1ex}{\textheight}}%WIDTH-LIMITED BIG WEDGE
  }{\textheight}% 
}{0.5ex}}%
\stackon[1pt]{#1}{\tmpbox}%
}

\newcommand{\ie}{\emph{i.e.}\@ifnextchar.{\!\@gobble}{}}
\newcommand{\eg}{\emph{e.g.}\@ifnextchar.{\!\@gobble}{}}
\newif\ifreview
\reviewfalse

\ifreview
	\usepackage{lineno}
	\renewcommand\thelinenumber{\color[rgb]{0.2,0.5,0.8}\normalfont\sffamily\scriptsize\arabic{linenumber}\color[rgb]{0,0,0}}
	\renewcommand\makeLineNumber {\hss\thelinenumber\ \hspace{6mm} \rlap{\hskip\textwidth\ \hspace{6.5mm}\thelinenumber}} 
	\linenumbers
\fi

\begin{document}

%%%%%%%%%%%%%%%%%%%%% Add submission id, track, and title. %%%%%%%%%%%%%%%%%%%%%

% TODO: Please insert your submission number here
\def\SubNumber{29}

% TODO: Please uncomment the track this paper will be submitted to, comment all other lines
\def\GCPRTrack{Main Track}
%\def\GCPRTrack{Special Track: Pattern recognition in the life and natural sciences}
%\def\GCPRTrack{Special Track: Photogrammetry and remote sensing}
%\def\GCPRTrack{Special Track: Computer vision systems and applications}
%\def\GCPRTrack{Young Researcher's Forum}
%\def\GCPRTrack{Fast Review Track}
%\def\GCPRTrack{Extended Abstract}

% TODO: Replace with your title
\title{Ordered Diffusion for 3D Human Registration}
% You can use \thanks for acknowledgment. Do not add any acknowledgment to the draft 
% version that is used for the review process.  
%\title{Title\thanks{XXX}}

\ifreview
	% ANONYMOUS SUBMISSION FOR REVIEW
	% DO NOT MODIFY these for the draft version that is used for the review process.
	\titlerunning{GCPR 2026 Submission \SubNumber{}. CONFIDENTIAL REVIEW COPY.}
	\authorrunning{GCPR 2026 Submission \SubNumber{}. CONFIDENTIAL REVIEW COPY.}
	\author{GCPR 2026 - \GCPRTrack{}}
	\institute{Paper ID \SubNumber}
\else
	% CAMERA READY SUBMISSION
	%\titlerunning{Abbreviated paper title}
	% If the paper title is too long for the running head, you can set
	% an abbreviated paper title here

    \author{Mattia Masiero\inst{1}\orcidlink{0009-0001-1327-0849} \and
    Ilya A. Petrov\inst{1,2}\orcidlink{0000-0002-8900-1071} \and \\
    Daniel Cremers\inst{3,4}\orcidlink{0000-0002-3079-7984} \and 
    Gerard Pons-Moll\inst{1,2,5}\orcidlink{0000-0001-5115-7794} \and
    Riccardo Marin\thanks{Corresponding Author}\inst{3,4}\orcidlink{0000-0003-2392-4612}}
	
	\authorrunning{M. Masiero et al.}
	% First names are abbreviated in the running head.
	% If there are more than two authors, 'et al.' is used.
    \institute{
    $^{1}${\small University of T\"ubingen, Germany}\qquad
    $^{2}$ {\small T\"ubingen AI Center, Germany}\\
    $^{3}$ {\small Technical University of Munich, Germany}\\
    $^{4}$ {\small Munich Center for Machine Learning, Germany}\\
    $^{5}$ {\small Max Planck Institute for Informatics, Saarland Informatics Campus, Germany}\\
    \url{https://riccardomarin.github.io/odin/}}
\fi

\maketitle              % typeset the header of the contribution

\begin{abstract}

3D human registration has historically been treated as a regression task, assuming a unique ground-truth alignment exists between the template and an input point cloud. In reality, acquisition noise, occlusions, and unknown soft tissue dynamics introduce inherent ambiguity into human scans. Regression-based methods consequently converge to an average prediction, often failing to represent a plausible geometry. In our work, we embrace such uncertainty by modeling the registration as a distribution of alignments. We propose \methodname, which formulates registration as a 3D diffusion process that generates a point cloud aligned with the target geometry while preserving template semantics through consistent point ordering. To achieve this, \methodname~relies on global, local, and positional conditioning, guiding each point to its correct location. Our experiments demonstrate that such a generative formulation not only outperforms its regression-based baseline, but also establishes a new state of the art, surpassing highly engineered methods while reducing the registration time by two-thirds.
\end{abstract}

%-------------------------------------------------------------------------
\section{Introduction}
\label{sec:introduction}

Finding correspondences between 3D objects is a cornerstone of countless downstream applications: from feature transfer to supervised learning; from camera estimation to shape editing. Among other classes, humans have attracted significant attention due to their practical applications. But such an interest does not come without challenges: finding dense semantic meaning in noisy, unstructured observations that are also subject to non-rigid deformations is particularly complicated, especially given the highly articulated nature of human bodies. A well-established strategy to tackle this task is to rely on a common template that provides an underlying structure and to solve for its alignment with the input point cloud.  Solving for such a deformation is commonly referred to as \textit{registration}.

Historically, registration has been formulated as a regression-based task, assuming that, given an input point cloud, only a single template alignment correctly fits it. But such simplification hides the uncertainty of this task -- an uncertainty that already arises from the acquisition, since sensors are noisy and occlusions can occur even with tens of RGB-D cameras. This ambiguity is further exacerbated when cameras are scarce and the environment is uncontrolled. However, even with complete observations, determining the semantic location of a specific body point can only be done with a certain degree of confidence due to the elasticity of soft tissues and body composition~\cite{bogo2017dynamic,pons2015dyna}, which can differ greatly between subjects. Reducing the task to a single prediction is not appropriate and often leads the method to make an average guess, which matches no single plausible hypothesis. 

These observations led us to challenge such a classical perspective. Our goal is to reframe registration as a generative task in which an input point cloud is associated with a distribution over plausible alignments. To do so, we build on diffusion models, phrasing the task as recovering a template from noise. While previous work has used diffusion models for 3D point cloud generation, it has mainly focused on \textit{reconstructing} the shape geometry from images, without accounting for the semantic meaning of the points. To achieve a denoising procedure that respects a pointwise semantic meaning, we present \textbf{\methodname}, the first \textbf{O}rdered \textbf{Di}ffusio\textbf{n} approach for 3D human registration. Since obtaining meaningful 2D images for colorless point clouds (especially when they are partial) is challenging, we propose a purely 3D-based conditioning with a balanced combination of positional, local, and global spatial information: a global latent code representing the entire target scan; per-point scan features assigned to the denoising template points; and a sinusoidal positional encoding. The diffusion process autonomously balances these conditioning sources throughout the denoising process without requiring a conditioning schedule, making our formulation both straightforward and generalizable. \methodname~outperforms the same backbone network trained for single-step regression, as well as other state-of-the-art approaches that rely on neural multi-step displacement fields, significantly reducing the error while taking even less than a third of the inference time. Also, we demonstrate our capabilities in cases where limbs are missing or shape observations are partial. In summary, our contributions are:
\begin{itemize}
    \item We propose \methodname, the first approach to tackle 3D human body registration as a diffusion-based process, opening a new research direction orthogonal to previous work.
    \item We deploy a purely 3D-based conditioning scheme that combines global, local, and positional information, achieving a point cloud denoising process that outputs points in an order coherent with a template.
    \item We achieve state-of-the-art performance on full-body and partial 3D human registration; we surpass well-established, engineered regression baselines, demonstrating the practical usefulness of our approach.
\end{itemize}
Our findings suggest that the proposed generative formulation is a promising new research direction, which we aim to foster by releasing and maintaining our code and pretrained models on the dedicated project page.

%-------------------------------------------------------------------------
\section{Related Work}
\label{sec:related_work}

\subsection{Correspondence and Registration}
The problem of aligning 3D data dates back to the origins of Computer Vision. Here, we review some key works relevant to our discussion, but for a comprehensive discussion, we refer the reader to~\cite{deng2022survey,zhuravlev2026non,heuschling2026establishing}.
Solving for correspondence can be phrased as a complex combinatorial problem~\cite{windheuser2011geometricaly}. This formulation has recently been rediscovered~\cite{roetzer2022scalable,roetzer2025fast}, although it remains slow and demanding of high input quality. More efficiently, the problem can be reformulated in the functional domain~\cite{ovsjanikov2012functional}, which admits extensions for volumes~\cite{maggioli2026volumetric}, point clouds~\cite{donati2020deep,marin2020correspondence}, and partial data~\cite{attaiki2021dpfm,ehm2024partial,ehm2025echomatch}, also thanks to significant advances in 3D feature extractor backbones~\cite{sharp2022diffusionnet,Gao_2026_CVPR}. However, their reliance on local geometry hinders their applicability in real point clouds coming from noisy acquisition pipelines. A common approach to address noise is to perform registration, in which a common template is aligned with the target shapes. The most famous approach is the seminal ICP~\cite{besl_method_1992}, which iterates between finding correspondences and applying a deformation. Its formulation has been revised multiple times to also handle non-rigid cases~\cite{amberg_optimal_2007,li_sumner_2008} and to incorporate data priors~\cite{bogo2016keep,pavlakos2019expressive,marin2020farm,marin2019high}. Recently, foundational and diffusion features have been used to find correspondences in 2D images~\cite{zhang2023tale,hartwig2025geco,cuttano2026marco} and then to match 3D shapes~\cite{dutt2024diffusion,ehm2026teaching}. More interestingly, we also observed advances in 3D, where diffusion models can generate plausible rigid roto-translations~\cite{An2024DiffusionTF} or doubly stochastic matrices~\cite{wu2023diff,wu2024diff} and are being integrated into functional map pipelines~\cite{zhuravlev2025denoising,pierson2025diffumatch}. Arguably, the works most closely related to ours are  DiffSurf \cite{yoshiyasu2024diffsurf}, which diffuses template mesh vertices undconditionally or from a given skeleton; SRIF~\cite{sun2024srif}, which relies on multiple renderings, an image deformation backbone~\cite{zhang2024diffmorpher}, and Gaussian splatting~\cite{kerbl20233d}; FUSE~\cite{olearo2025fuse}, an optimization-based approach that requires multiple ground-truth landmarks. To the best of our knowledge, no methods have attempted to use diffusion to solve registration of unordered input colorless point clouds solely from the input unordered 3D scans, without the use of extra visual or semantical priors. 

\subsection{3D Human Registration}
Initially, 3D human registration was tackled through intensive manual labor, complex annotation pipelines, and visual context~\cite{bogo_faust_2014,bogo2017dynamic}. Ambitiously, our work follows the path of automatic methods that rely solely on the 3D geometric information. A milestone in deep learning for 3D registration was the seminal 3D-Coded~\cite{Groueix_2018_ECCV}, although its encoder-decoder architecture proved insufficient for scaling. TransMatch~\cite{trappolini2021shape}, which uses a powerful attention-based architecture, also faces a similar bottleneck. IPNet~\cite{bhatnagar_combining_2020} instead relies on implicit representations and simple part-based chamfer optimization, which can struggle with challenging inputs. Relying on locality priors, PTF~\cite{wang2021locally} predicts SMPL parameters using local feature extraction, which is not easily generalizable to partial observations. Previous work has also explored learning without ground-truth supervision: LoopReg~\cite{bhatnagar2020loopreg} proposes a self-supervised schema for full shapes, while the recent SS-HMR~\cite{su2025self} proposes a multi-initialization self-improving loop tailored to partial shapes. Both show promising performance, but mainly focus on the same distribution observed during training. With the advent of neural fields, registration has been formulated as learning a displacement field. While solving registration in a single deformation step requires strong regularizations~\cite{sundararaman2022reduced}, 
LVD~\cite{corona_learned_2022} leverages a neural field that is iteratively queried to guide template points towards their target locations. Yet, such approach can often lead to predictions that do not align with the target surface. NICP~\cite{marin_nicp_2025} extended LVD to provide geometric grounding, promoting template points to lie on the target surface via a test-time adaptation task. To date, NICP is the state of the art for 3D unclothed human registration, and we consider this our most relevant baseline. It is also worth noting that a recent line of work considers the registration of humans under clothing~\cite{li2025etch,li2026etch}, which we recognize as an interesting setting for future application of our generative formulation.

\subsection{Diffusion for 3D}
In 2015, Sohl-Dickstein et al.~\cite{sohl2015deep} first introduced diffusion models showing that data distributions can be recovered by reversing a diffusion-like stochastic process, though at high computational cost. After the efficiency gains introduced by~\cite{ho_denoising_2020}, diffusion models have spread to almost every domain, including 3D data. An early work in this direction is the one from Luo et al. ~\cite{Luo_2021_CVPR}, which applies diffusion on 3D point clouds, conditioning the process by a global shape latent code extracted with PointNet~\cite{qi2017pointnet}. LION~\cite{vahdat2022lion} addresses the task of 3D shape reconstruction from point clouds
using diffusion in a hierarchical latent space representation. It supports both conditional and unconditional generation. PC$^2$~\cite{melas2023pc2} addresses category-level point cloud reconstruction from a single RGB image. It conditions the diffusion process by projecting the diffusion point cloud onto an image feature volume at each step. GECCO~\cite{tyszkiewicz2023gecco} uses a permutation-invariant set transformer~\cite{lee2019set} and follows a similar approach to PC$^2$, but replaces the rasterization-based projection with simpler bilinear interpolation. MeshDiffusion~\cite{liu2023meshdiffusion} operates on 3D meshes, using tetrahedral grids and differentiable rendering. An interesting idea is presented in PDT~\cite{wang2025pdt}, where an input mesh or point cloud is denoised toward specific geometric salient distributions (e.g., vertices, lines, or joints). Such semantic information could help recover structures in shape collections, but it still does not help recover clear semantics.
Closer to the human domain, diffusion has been used to generate parametric bodies jointly with objects~\cite{petrov2024tridi}, whose transformer backbone we adapt, and to recover full-body interactions from sparse egocentric observations~\cite{petrov2026echo}. In these works, however, the point semantics are imposed by the parametric model, whereas we recover them directly from an unstructured observation. 
%-------------------------------------------------------------------------
\section{Background} 
\label{sec:background}
\subsection{SMPL Body Model}
SMPL~\cite{loper2023smpl} is the de facto standard in 3D human body modeling. SMPL consists of a mesh template of $N=6890$ vertices that can be controlled by shape parameters $\boldsymbol{\beta} \in \mathbb{R}^{|\boldsymbol{\beta}|}$ and pose parameters $\boldsymbol{\theta} \in \mathbb{R}^{3K}$ where $K=24$ is the number of joints. The first are the coefficients of a PCA space and are generally restricted to the first $10$ dimensions, while the second encode the disposition of the body articulations, where the joints are arranged in a kinematic tree rooted at the pelvis, giving $|\boldsymbol{\theta}| = 72$. Each joint's local rotation relative to its parent is represented as a rotation vector (axis-angle), with the global body orientation encoded in the root joint. Given pose and shape parameters, the final mesh vertices $\mathbf{M}(\boldsymbol{\theta}, \boldsymbol{\beta}) \in \mathbb{R}^{N \times 3}$ are obtained via Linear Blend Skinning (LBS). Joint locations are regressed from the shaped template, after which the forward kinematic chain computes a rigid transformation $\mathbf{G}_k(\boldsymbol{\theta}, \mathbf{J})$ for each joint $k$. Each vertex $i$ (in homogeneous coordinates) is deformed as a weighted sum of joint transformations:
\begin{equation}
    \mathbf{M}_i(\boldsymbol{\theta}, \boldsymbol{\beta}) = \sum_{k=1}^{K} w_{k,i} \, \mathbf{G}_k\!\left(\boldsymbol{\theta}, \mathbf{J}(\boldsymbol{\beta})\right) \left(\bar{\mathbf{T}}_i(\boldsymbol{\beta}) + B_P(\boldsymbol{\theta})_i\right),
\end{equation}
where $\mathbf{J}(\boldsymbol{\beta})$  are the regressed joints, $w_{k,i}$ are the pre-learned blend skinning weights governing the influence of joint $k$ on vertex $i$, $\bar{\mathbf{T}}_i(\boldsymbol{\beta})$ is the template mesh in neutral pose with the identity defined by the shape parameters, and $B_P(\boldsymbol{\theta})_i$ the pose corrective term. This formulation makes the SMPL model fully differentiable with respect to both $\boldsymbol{\theta}$ and $\boldsymbol{\beta}$. Our method aims to diffuse 3D points in an order consistent with the SMPL template, using the SMPL parameters just to refine the model output.

\subsection{Diffusion Models}
\paragraph{Training.} Diffusion models~\cite{sohl2015deep,ho_denoising_2020} are generative models that produce data by progressively denoising a sample drawn from a standard Gaussian distribution. Training starts with a clean data sample \( \mathbf{x}_0 \) that is corrupted according to the Gaussian forward transition \( t = 1, \dots, T \):
\begin{equation}
q(\mathbf{x}_t \mid \mathbf{x}_{t-1}) = \mathcal{N}\left(\mathbf{x}_t; \sqrt{1 - \gamma_t} \, \mathbf{x}_{t-1}, \, \gamma_t \mathbf{I} \right),
\end{equation}
where \(\gamma_t \in (0,1)\) is determined by a variance schedule and
controls the amount of added noise. For efficient training, the marginal distribution at any timestep \( t \) can be expressed in closed form as
\begin{equation}
q(\mathbf{x}_t \mid \mathbf{x}_0) = \mathcal{N}\left(\mathbf{x}_t; \sqrt{\bar{\alpha}_t} \, \mathbf{x}_0, \, (1 - \bar{\alpha}_t)\mathbf{I} \right),
\end{equation}
where \( \alpha_t = 1 - \gamma_t \) and \( \bar{\alpha}_t = \prod_{s=1}^t \alpha_s \). The denoiser \( D_\phi(\mathbf{x}_t, t) \) is implemented as a neural network that predicts the noise \( \boldsymbol{\epsilon} \)~\cite{ho_denoising_2020} and is trained to minimize

\begin{equation}
\mathcal{L}
=
\mathbb{E}_{\boldsymbol{\epsilon} \sim \mathcal{N}(0,\mathbf{I}),\, t \sim \mathcal{U}(\{1,\ldots,T\})}
\left[
\left\|
\boldsymbol{\epsilon}
-
D_\phi(\mathbf{x}_t, t)
\right\|_2^2
\right].
\end{equation}

\paragraph{Backward Process.}
Once the network \(D_\phi\) has been trained, generation starts with a noise sample
\(\mathbf{x}_T \sim \mathcal{N}(\mathbf{0},\mathbf{I})\), which is
iteratively denoised through the learned reverse transitions
\begin{equation}
p_\phi(\mathbf{x}_{t-1}\mid\mathbf{x}_t)
=
\mathcal{N}\left(
\mathbf{x}_{t-1};
\boldsymbol{\mu}_\phi(\mathbf{x}_t,t),
\sigma_t^2\mathbf{I}
\right),
\end{equation}
where $\sigma_t^2$ is the posterior variance. The predicted noise \(\widetilde{\boldsymbol{\epsilon}} = D_\phi(\mathbf{x}_t,t)\) is used to estimate the clean sample as
\begin{equation}
\tilde{\mathbf{x}}_0
=
\frac{
\mathbf{x}_t
-
\sqrt{1-\bar{\alpha}_t}\,
\widetilde{\boldsymbol{\epsilon}}
}{
\sqrt{\bar{\alpha}_t}
}.
\end{equation}

Since the forward process is linear Gaussian, the posterior
\(q(\mathbf{x}_{t-1}\mid\mathbf{x}_t,\mathbf{x}_0)\) is Gaussian with
closed-form mean
\begin{equation}
\widetilde{\boldsymbol{\mu}}_t(\mathbf{x}_t,\mathbf{x}_0)
=
\frac{\sqrt{\bar{\alpha}_{t-1}}\gamma_t}
     {1-\bar{\alpha}_t}\mathbf{x}_0
+
\frac{\sqrt{\alpha_t}(1-\bar{\alpha}_{t-1})}
     {1-\bar{\alpha}_t}\mathbf{x}_t.
\end{equation}
Substituting \(\tilde{\mathbf{x}}_0\) for \(\mathbf{x}_0\) in the posterior mean yields the estimated reverse-process mean
\(\boldsymbol{\mu}_\phi(\mathbf{x}_t,t)
=
\widetilde{\boldsymbol{\mu}}_t
\left(\mathbf{x}_t,\tilde{\mathbf{x}}_0\right)\).
%Here, \(\sigma_t^2\) is set according to the standard DDPM fixed variance schedule.
Iterating the reverse transitions produces the final denoised sample
\(\mathbf{x}_0\).

\paragraph{Point Cloud Generation via Diffusion.} The PC$^2$ formulation~\cite{melas2023pc2} proposes a diffusion model for generating 3D point clouds conditioned on a 2D image. An image encoder first extracts a feature volume \(\mathbf{F}\). Each point in \(\mathbf{x}_t\) is projected onto the image plane to retrieve a corresponding feature from \(\mathbf{F}\), yielding the conditioning features \(\mathbf{f}\). The denoiser predicts the diffusion noise as
\begin{equation}
\tilde{\boldsymbol{\epsilon}} = D_{\text{PC}^2}(\mathbf{x}_t, \mathbf{f}, t)
\end{equation}
Like PC$^2$, our method uses global and local conditioning. However, we condition on structured 3D features instead of 2D image features, making the method applicable to point clouds, including partial scans. In addition, our method preserves the semantic ordering of the output points, maintaining correspondence with the SMPL template.
%-------------------------------------------------------------------------
\section{Ordered Diffusion (ODin)} 
\label{sec:method}
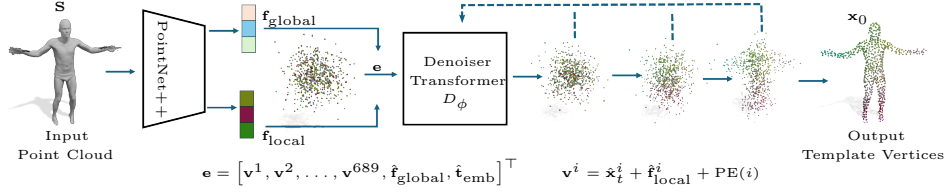
\begin{figure*}[t!]
    \centering
    \begin{overpic}[width=\linewidth]{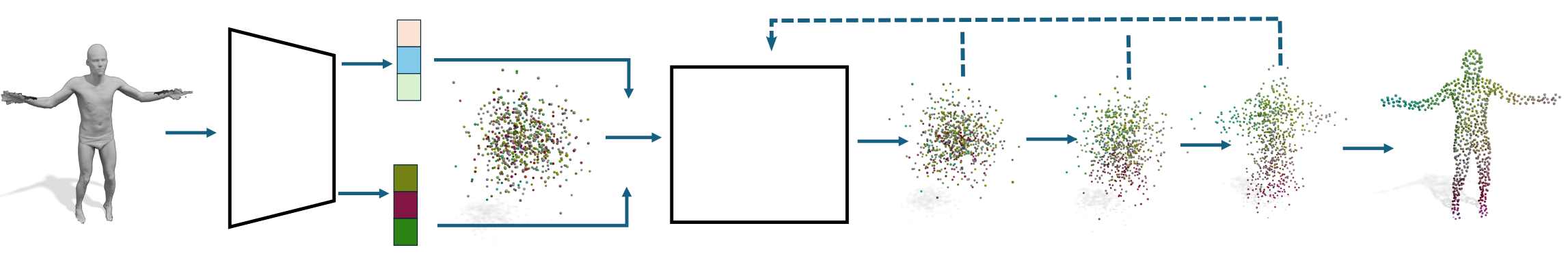}
    \tiny
        \put(5, 15){$\mathbf{S}$}
        \put(4, 1){Input}
        \put(1, -1){Point Cloud}

        \put(16, 14){\rotatebox{-90}{PointNet++}}

        \put(27.5, 14){$\mathbf{f}_{\text{global}}$}
        \put(27.5, 1){$\mathbf{f}_{\text{local}}$}

        \put(39.5, 8.5){$\mathbf{e}$}

        \put(45, 9){Denoiser}
        \put(43.5, 7){Transformer}
        \put(47, 5){$D_\phi$}

        \put(91, 1){Output}
        \put(86, -1){Template Vertices}

        \put(21, -3){$\mathbf{e} = \left[ \mathbf{v}^1, \mathbf{v}^2, \dots, \mathbf{v}^{689}, \hat{\mathbf{f}}_\text{global}, \hat{\mathbf{t}}_\text{emb} \right]^\top$ }

         \put(60, -3){$\mathbf{v}^i = \hat{\mathbf{x}}_t^{i} + \hat{\mathbf{f}}_\text{local}^{i} + \text{PE}(i)$}

        \put(91,14){$\mathbf{x}_0$}
    \end{overpic}
    \vspace{0.01cm}
    
    \caption{\textbf{O}rdered \textbf{Di}ffusio\textbf{n} for 3D human registration. Given an input point cloud (depicted as a mesh for visualization), we first extract global and per-point local features using a PointNet++ backbone. Then, a denoising transformer learns to recover an ordered point cloud from Gaussian noise. At each step, the local features $\mathbf{f}_{\text{local}}$ are assigned to the diffusion points via nearest neighbors. Together with the global feature $\mathbf{f}_{\text{global}}$, they condition the diffusion process on the input scan. The output is a point cloud that fits the input geometry, with points ordered according to a template's (SMPL) ordering.}
    \label{fig:pipeline}
\end{figure*}

\paragraph{Overview.}
In the following, we present Ordered Diffusion (ODin), our approach for aligning a template with an input point cloud $\target \in \mathbb{R}^{P\times3}$ of $P$ points using a diffusion process. We first formulate a three-level conditioning to guide the denoised points toward the target geometry while maintaining the SMPL template ordering (Sec.~\ref{sec:features}). Then, we explain our denoising process at inference and how we use the network prediction to recover an aligned SMPL model faster than previous approaches (Sec.~\ref{sec:inference}). A depiction of our method can be seen in Fig.~\ref{fig:pipeline}.

\subsection{Conditioning}
\label{sec:features}
\paragraph{Intuition.} How to structure a purely 3D conditioning to inform the diffusion of the desired geometry, while maintaining a point order coherent with the template semantics? Our intuition is to combine three levels of information: global and per-point features from a 3D feature extractor, and a positional encoding applied to the diffused 3D points. 
Global features describe the overall shape, while local features capture fine-grained pose details. The positional encoding enables the network to learn the role of each point based on its index.

\paragraph{Positional Encoding.} The denoising model $\denoiser$  takes as input perturbed template points $\mathbf{x}_t^{i}$ and predicts their noise $\boldsymbol{\epsilon}$. Intuitively, the noise here represents a 3D offset that should move every template point toward the correct position. However, without knowing the identity of each point, such an operation is impossible because the transformer is permutation-equivariant. To this end, we first decorate every point with the canonical sinusoidal positional encoding  $\text{PE}(i)$ introduced in~\cite{vaswani2017attention}, thus associating it with its identity on the template shape:

%\begin{equation}
%\begin{aligned}
%\mathrm{PE}(i,2k)
%&=
%\sin\left(\frac{i}{10000^{2k/d}}\right),\\
%\mathrm{PE}(i,2k+1)
%&=
%\cos\left(\frac{i}{10000^{2k/d}}\right),
%\end{aligned}
%\qquad
%k=0,1,\ldots,\frac{d}{2}-1,
%\end{equation}

\begin{equation}
\begin{aligned}
\mathrm{PE}(i,2k) &= \sin(i/10000^{2k/d}),\\
\mathrm{PE}(i,2k+1) &= \cos(i/10000^{2k/d}),
\end{aligned}
\qquad
k=0,\ldots,d/2-1,
\end{equation}
where $i$ is the ordinal number of the vertex in the template, and $d=1024$.

\paragraph{Global and Local Features.} To condition the diffusion on the input scan, we use information from both its global and local structure. We pass the target scan points $\target$ through a PointNet++ encoder-decoder~\cite{qi2017pointnet++} and extract global features $\textbf{f}_{\text{global}} \in \mathbb{R}^{256}$ at the bottleneck layer, using a single PointNet++ centroid for the entire point cloud. We also extract per-point local features $\textbf{f}_{\text{local}} \in \mathbb{R}^{P \times128}$ after the final feature-propagation layer. However, while global features can be passed to the denoiser as an additional token, it is not immediately clear how to incorporate local features into a 3D diffusion process, as they must be assigned to the noisy points. Our idea is that, since both the scan and the Gaussian noise are centered at zero and of appropriate scale, at every diffusion step, for every noisy point $\mathbf{x}_t^{i}$, we find the closest point on the input scan:
\begin{equation}
\mathbf{s}^{*}:= \arg \min \limits_{\mathbf{s} \in \target} \|\mathbf{x}_t^{i} - \mathbf{s}\|_2.
\end{equation}
Hence, we assign the local features $\mathbf{f}^i_{\text{local}}$ of the corresponding input scan point $\mathbf{s}^{*}$ to the noised 3D point $\mathbf{x}_t^{i}$. This step might look natural, but it is also counterintuitive, since for $t \approx T$ (i.e., when the diffused points are almost pure noise) such a pairing is basically random. However, we found it effective, suggesting that the transformer can leverage different conditioning sources at different timesteps. In particular, when $t \approx 1$, the diffused points are more closely aligned with the target shape, so nearest-neighbor feature assignments provide a more informative local conditioning signal.

\paragraph{ODin Input Tokens.} 
In summary, to build the input to our diffusion process, we first extract a global feature $\mathbf{f}_{\text{global}}$ and per-point local features $\mathbf{f}_{\text{local}}$ from the scan by performing a forward pass through the PointNet++ backbone. We assign local features to each diffusion point and apply positional embeddings. Then, the transformer inputs are mapped to a common $\mathbb{R}^{d}$ embedding space via MLPs. This yields $1024$-dimensional tokens $\hat{\mathbf{x}}_t^{i}$ for the noised diffusion points, $\hat{\mathbf{f}}_\text{global}$ and $\hat{\mathbf{f}}_\text{local}^{i}$ for global and local scan features respectively, and $\hat{\mathbf{t}}_\text{emb}$ for the timestep embedding. Each point of the diffused ordered point cloud is represented as:
\begin{equation}
\mathbf{v}^i = \hat{\mathbf{x}}_t^{i} + \hat{\mathbf{f}}_\text{local}^{i} + \text{PE}(i).
\end{equation}
Following~\cite{marin_nicp_2025}, we also consider a subset of $689$ SMPL vertices in our implementation, which suffices to fit the full SMPL model afterward, while keeping the network complexity reasonable.
Hence, the input to self-attention is:
\begin{equation}
\mathbf{e} = \left[ \mathbf{v}^1, \mathbf{v}^2, \dots, \mathbf{v}^{689}, \hat{\mathbf{f}}_\text{global}, \hat{\mathbf{t}}_\text{emb} \right]^\top \in \mathbb{R}^{691 \times d}.
\end{equation}

\paragraph{Training.}
To train ODin, we use a set of human 3D point cloud scans with associated ground-truth template alignment. For each scan, we extract the corresponding global feature \(\mathbf{f}_{\text{global}}\) and local features \(\mathbf{f}_{\text{local}}\). Then, we compose batches where the ground-truth registration is corrupted by a random amount of noise parametrized by $t$, and we train the model using the following loss:

\begin{equation}
\mathcal{L}_{\text{diff}}
=
\mathbb{E}_{\boldsymbol{\epsilon} \sim \mathcal{N}(0,\mathbf{I}),\, t \sim \mathcal{U}(\{1,\ldots,T\})}
\left[
\left\|
\boldsymbol{\epsilon}
-
D_\phi(\mathbf{x}_t, \mathbf{f}_{\text{global}}, \mathbf{f}_{\text{local}}, t)
\right\|_2^2
\right].
\end{equation}
The PointNet++ and the transformer $D_{\phi}$ modules are trained end-to-end.

\subsection{Inference}
\label{sec:inference}
Given an input scan, we sample 4096 points and use the same conditioning as in training. At each step, the denoising model predicts the full noise, which is partially used to reduce the input Gaussian noise. This process is repeated until the final output is obtained. We follow the DDPM~\cite{ho_denoising_2020} sampling schedule. 

\paragraph{SMPL Fitting.}
After generating the ordered SMPL vertices, as in previous works~\cite{corona_learned_2022,marin_nicp_2025}, we use the network's prediction to recover the SMPL instance that best matches the diffusion output. Unlike previous methods, we employ \textit{SMPLfitter}~\cite{sarandi24nlf}, which is significantly faster than standard gradient-based optimization without sacrificing performance. We further refine the SMPL instance by optimizing:

\begin{equation}
\mathcal{L}_{\text{fit}}(\boldsymbol{\theta}, \boldsymbol{\beta}) = \text{CD}(\target, \mathbf{M}(\boldsymbol{\theta}, \boldsymbol{\beta}))+ w_{vp}\text{VP}(\boldsymbol{\theta}) + w_\beta \|\boldsymbol{\beta}\|_2^2
,
\end{equation}
where CD is chamfer distance, VP the VPoser regularization~\cite{pavlakos2019expressive}, $w_{vp}=10^{-3}$, and  $w_\beta=0.2$. Throughout our text, we refer to this last refinement as \textit{chamfer optimization}. When shapes are complete, we consider the bidirectional chamfer distance, while in the case of partiality, we use the unidirectional one.
%-------------------------------------------------------------------------
\section{Experiments}
\label{sec:experiments}
\begin{figure}[t!]
    \centering
    \begin{overpic}[width=\linewidth, trim=0 0 0 0, clip]{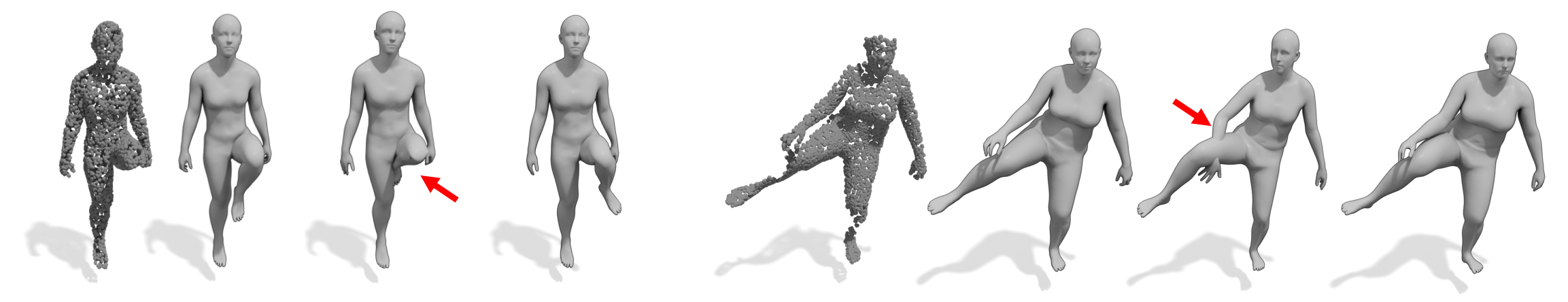}
    \footnotesize
        \put(3, -1.5){Input}
        \put(12, -1.5){GT}
        \put(20, -1.5){NICP~\cite{marin_nicp_2025}}
        \put(34, -1.5){\textbf{\methodname}}

        \put(49, -1.5){Input}
        \put(62, -1.5){GT}
        \put(72, -1.5){NICP~\cite{marin_nicp_2025}}
        \put(87, -1.5){\textbf{\methodname}}
        
    \end{overpic}
     % \vspace{0.01cm}
    
    \caption{Comparison of the methods' outputs in the missing-limb and partial-view settings. All results are pre-chamfer optimization on the DFAUST test set with partiality (missing left leg and a partial view). In the presence of partiality, NICP often predicts unrealistic penetrations. \methodname~provides better predictions, resulting in more accurate registration.}
    \label{fig:results}
\end{figure}
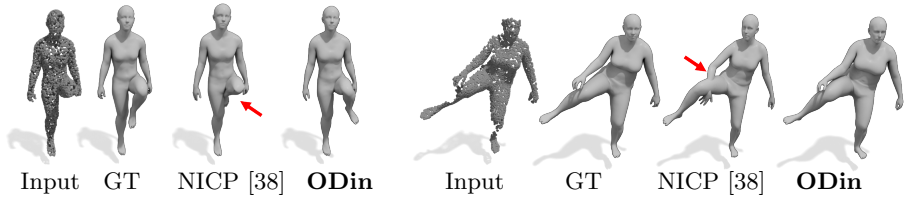

\subsection{Setup}
\paragraph{Implementation.} The denoiser is trained end-to-end together with the point cloud feature extractor for 100{,}000 update steps using a batch size of 16. We use a cosine learning-rate schedule with a peak learning rate of 0.0005 for the denoiser transformer and 0.005 for the PointNet++ feature extractor. The peak learning rate is reached after 2000 warm-up steps. We optimize the model using AdamW. The diffusion process consists of 1000 timesteps, while at inference, we use 100 DDPM sampling steps. We use a linear noise schedule with $\gamma_1 = 1 \times 10^{-5}$ and $\gamma_T = 8 \times 10^{-3}$, controlling the initial and final noise levels, respectively. 
%For sampling, we employ a DDPM scheduler to follow the backward diffusion process. 
We run our experiments on a workstation with an NVIDIA GeForce RTX 3080 Ti.

\paragraph{Datasets.} To conduct our experiments, we train our network on two different datasets. First, we use DFAUST~\cite{bogo2017dynamic}, a 4D dataset of real scans with roughly 150k vertices each, totaling around 40k frames, each paired with a ground-truth SMPL mesh.
Then, to demonstrate the scalability of our method, we also train it on AMASS~\cite{mahmood2019amass} (excluding DFAUST subset), a large dataset of human poses comprising 300 subjects and 11k motions. For each AMASS frame we use, we sample points from the corresponding SMPL mesh and perturb the resulting point cloud to simulate scan noise. Since temporal sequences are redundant, we subsample them, yielding 4k training shapes for DFAUST and 170k for AMASS. For DFAUST, we use the same train/test split as ArtEq~\cite{feng2023generalizing}, resulting in 313 test samples, which is comparable in scale to standard human body registration benchmarks~\cite{bogo_faust_2014,melzi2019shrec} and literature standards~\cite{feng2023generalizing,marin_nicp_2025}. For testing, we also consider FAUST~\cite{bogo_faust_2014}, which provides $100$ real scans with annotated ground truth. Finally, we provide results on shapes with one peripheral limb (the outward half) removed (DFAUST$_{ml}$). In the supplementary materials, we also report qualitative results from SHREC19~\cite{melzi2019shrec} and more missing-limb completion examples. All tables report the mean vertex-to-vertex error, averaged over all template vertices, in centimeters.

\paragraph{Baselines.} We consider NICP~\cite{marin_nicp_2025} as the main competitor to our method. NICP is the current state of the art in unclothed human body registration, demonstrating robustness to partiality and noise. There is also an interesting analogy to our approach: the backbone of NICP iteratively computes the registration by updating the template vertices' positions by small displacements. Our diffusion process is similar: we also progressively denoise template points toward the target surface. However, NICP is deterministic and relies on test-time adaptation via fine-tuning. Instead, our approach models uncertainty directly within the method, allowing for a distribution of possible solutions. We also compare with ArtEq~\cite{feng2023generalizing}, a representative single-pass registration method. ArtEq relies on localized SE(3)-equivariant/invariant features and generalizes to difficult poses, although its theoretical assumptions are unrealistic for real-world point clouds.

\subsection{Results}

% new results
\begin{table}[t!]
    \centering
    % Prima Tabella (Sinistra)
    \begin{minipage}{0.57\textwidth}
        \centering
        \begin{tabular}{lcc}
            \toprule
            & \makecell{DFAUST \\ err (cm) $\downarrow$} & \makecell{DFAUST$_{ml}$ \\ err (cm) $\downarrow$} \\
            \midrule
            \textbf{\methodname} & \textbf{1.15/0.79} & \textbf{4.80/3.87} \\
            NICP~\cite{marin_nicp_2025} & 1.47/1.00 & 5.75/5.38 \\
            \midrule
            \textbf{\methodname}$_{ml}$ & \textbf{1.14/0.78} & \textbf{2.22/2.14} \\
            NICP$_{ml}$~\cite{marin_nicp_2025} & 1.49/0.99 & 2.56/2.59 \\
            \bottomrule
        \end{tabular}
        \caption{Methods trained on AMASS ($ml$ denotes the use of missing-limbs shapes at training time) and tested on DFAUST ($ml$ denotes that test data have missing limbs); results are without/with chamfer optimization.}
        \label{tab:DFAUST}
    \end{minipage}
    \hfill
    % Seconda Tabella (Destra)
    \begin{minipage}{0.4\textwidth}
        \centering
        \begin{tabular}{lcc}
            \toprule
            \multicolumn{3}{c}{\textbf{FAUST}} \\
            \midrule
            & err (cm) $\downarrow$ & time [s] $\downarrow$ \\
            \midrule
            NICP~\cite{marin_nicp_2025} & 1.48 & 35 \\
            \textbf{\methodname} & \textbf{1.31} & \textbf{9} \\
            \bottomrule
        \end{tabular}
        \caption{Comparison on FAUST of methods trained on AMASS. Results are after chamfer optimization. }
        \label{tab:v2v_methods_faust}
    \end{minipage}
\end{table}

\paragraph{Full Scans.} We start by analyzing our method's results on the full DFAUST shapes. In Tab. \ref{tab:DFAUST}, the first two rows compare our method with NICP, both trained on AMASS. For both methods, we report two numbers: the network's registration prediction and the result after the chamfer optimization. In the first column, we generally observe that on the DFAUST full-shapes test set, \methodname~outperforms NICP both before and after the chamfer optimization. Our inference yields a better local minimum, avoiding semantic errors such as penetrations. To give an intuition of the kinds of mistakes, we refer to Fig. \ref{fig:results}. On the left, the deterministic NICP provides an average guess for the missing left leg, resulting in an unrealistic registration. \methodname~provides a more reasonable pose and thus a better local minimum. On the right, we show a similar behavior on a simulated partial view. We report more discussion on this in the supplementary materials. In Tab.~\ref{tab:v2v_methods_faust}, we also report results for models trained on the AMASS dataset tested on the 100 FAUST scans with public ground truth. As with the DFAUST test sets, we see a similar advantage. In Tab. \ref{tab:v2v_methods_faust}, we also compare methods in terms of runtime, measuring the whole pipeline (from input sampling to chamfer optimization), showing that \methodname~runs significantly faster than NICP.

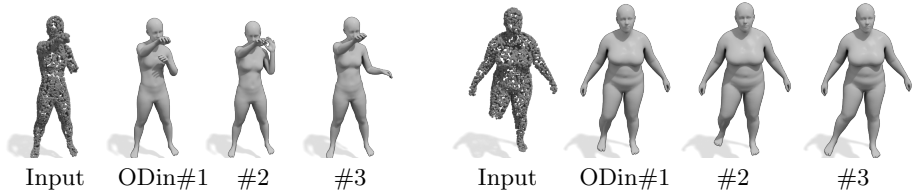
\begin{figure*}[t!]
    \centering
    \footnotesize
    \begin{overpic}[width=\linewidth, trim=0 20 0 0, clip]{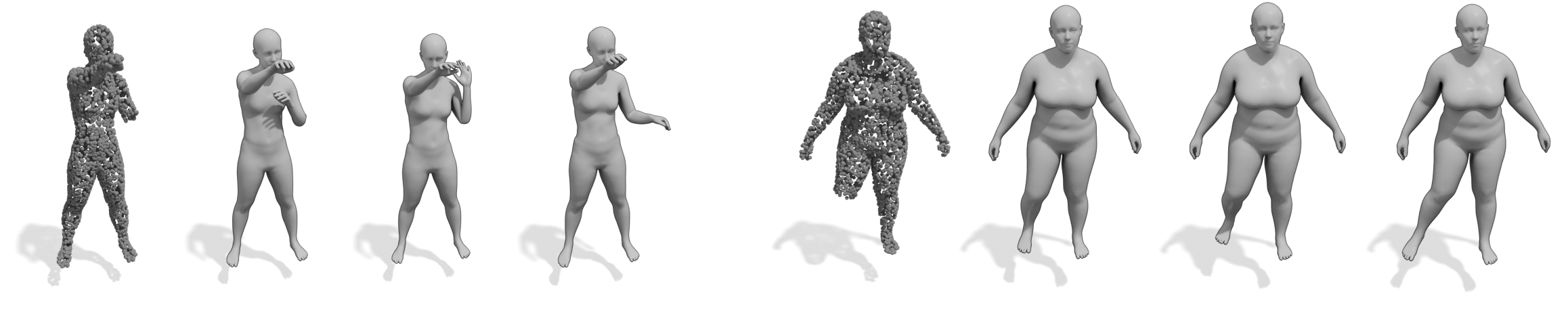}
        \put(3, -3){Input}
        \put(13, -3){\methodname \#1}
        \put(25, -3){ \#2}
        \put(35.5, -3){ \#3}

        \put(52, -3){Input}
        \put(63, -3){\methodname \#1}
        \put(77, -3){ \#2}
        \put(90, -3){ \#3}
        
    \end{overpic}
    \vspace{0.05cm}
    
    \caption{Results on partiality. Given a scan with missing observations, ODin can produce multiple plausible hypotheses for the missing part.}
    \label{fig:partiality}
\end{figure*}

\paragraph{Partial Scans.} In Tab. \ref{tab:DFAUST}, we also report results for the methods trained and tested on shapes with missing limbs (denoted as $ml$). 
This training regime still favors our method, both before and after the chamfer optimization. Intuitively, in the absence of a limb, the deterministic formulation of NICP predicts an average solution, often glued to the body, resulting in unrealistic poses with artifacts such as penetrations. In contrast, \methodname~samples from a distribution. The chosen solution might not align with the ground truth, but the pose is realistic and improves with regularization during chamfer optimization. We report examples of different predictions from our network in the presence of partiality in Fig.~\ref{fig:partiality}, showing that ODin captures the nuances of limb variability (e.g., the forearm has more degrees of freedom than the lower leg).

\subsection{Ablation}
For the ablation study, we train and test on DFAUST, since it is compact and provides real scans. We also include ArtEq~\cite{feng2023generalizing}, which uses the same DFAUST splits.

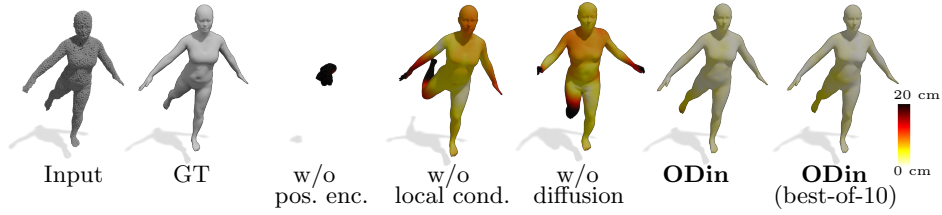
\begin{figure*}[t!]
    \centering
    \footnotesize
    \begin{overpic}[width=\linewidth]{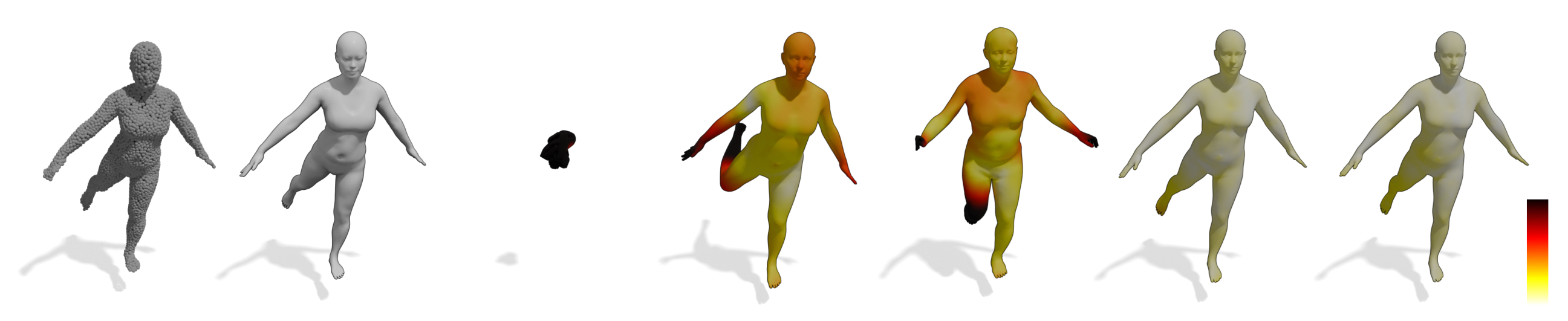}
        \put(5, -0.7){Input}

        \put(19, -0.7){GT}
        
        \put(32, -0.7){w/o}
        \put(30, -3){pos. enc.}
        
        \put(46.5, -0.7){w/o}
        \put(43, -3){local cond.}
        
        \put(60.5, -0.7){w/o}
        \put(58, -3){diffusion}
        
        \put(72, -0.7){\textbf{\methodname}}
        
        \put(87, -0.7){\textbf{\methodname}}
        \put(84, -3){(best-of-10)}

        \put(97, 0){\tiny 0 cm}
        \put(97, 8.3){\tiny 20 cm}

    \end{overpic}
    \vspace{0.05cm}
    
    \caption{Ablation study. Colors encode errors, with saturation at $20$ centimeters. Removing the positional encoding causes collapses, while without local conditioning, there are still significant misalignments. The same backbone in a deterministic setting fails to recover the correct pose. Our model also allows for multiple samplings (best-of-10).}
    \label{fig:ablation}
\end{figure*}

\begin{table}[t!]
\centering
\begin{tabular}{p{3cm}C{2cm}C{2cm}}
\toprule
\multicolumn{3}{c}{\textbf{DFAUST}} \\
\hline
Method & \makecell{w/o~cham \\ err (cm) $\downarrow$} &  \makecell{full-method \\ err (cm) $\downarrow$}\\
\hline
{\textbf{\methodname} (Single pred)} & 1.72 & \textbf{1.24}\\

{\textbf{\methodname} (best-of-10)} & \textbf{1.58} & \textbf{1.23}\\
\hline
{w/o pos. enc.} & 48.82 & 66.24\\
{w/o local cond.} & 16.68 & 6.77\\
{w/o diffusion} & 10.86 & 3.45\\
{ArtEq~\cite{feng2023generalizing}} & - & 3.62\\

\end{tabular}
\caption{\label{tab:ablation} Ablation study on methods trained and tested on DFAUST. We report the vertex-to-vertex error, before and after the chamfer optimization. All our design choices contribute significantly to reducing the error.}
\end{table}

\paragraph{Without Positional Encoding.}
Removing the positional encoding causes a collapse, since the transformer cannot learn the ordering of points and, in turn, fails to preserve the semantics of the template vertices. Consequently, the fitting to recover the full SMPL model results in catastrophic failure, as seen in Tab. \ref{tab:ablation} and Fig. \ref{fig:ablation}.

\paragraph{Without Local Conditioning.}
The model has to rely on a global feature descriptor of the scan and the positional encoding. The predictions are coarser, with less detail fidelity, highlighting the need for such localized information.

\paragraph{Without Diffusion (Deterministic).}
To demonstrate the usefulness of our probabilistic formulation, we also present results using the same transformer architecture, trained to predict per-point offsets from a fixed neutral SMPL template. We use the same conditioning information as \methodname~and attach the corresponding features from the PointNet++ backbone to each template point. We observe not only that the limbs are misplaced, but also that the posture appears inaccurate, probably because the deterministic network has converged on an average prediction and would require longer training to capture specific nuances.

\paragraph{Single Prediction vs. Best-of-10.} Our probabilistic formulation allows for navigating through multiple options. In this setting, we draw multiple samples from the diffusion model and select the prediction that minimizes the chamfer distance to the input scan. The best-of-10 formulation yields better predictions, whereas chamfer optimization obtains comparable results, indicating that the multiple samples lie within the same local minimum. For consistency with the prior literature, we based our experiments on the single-prediction approach, but drawing multiple hypotheses remains a flexible alternative to different computational budget needs. In both cases, our prediction improves over ArtEq~\cite{feng2023generalizing} by more than 50\%.

%-------------------------------------------------------------------------

\section{Conclusions}
\label{sec:conclusions}
In this work, we proposed the first diffusion-based pipeline for 3D human registration, opening a new research direction. \methodname~is designed around 3D-based conditioning with three levels of granularity, enabling the diffusion model to preserve the semantic meaning of the points. Our generative approach not only achieves state-of-the-art results in the full-body setting but is also applicable when information is missing, such as limbs or partial views.
Although our work focuses on humans, our design is general and can be extended to other classes, including those that do not enjoy a parametric template such as SMPL. We also see significant potential in incorporating the diffusion toolbox into the registration process, including score distillation sampling~\cite{poole2022dreamfusion}, guidance~\cite{ho2021classifierfree}, and different noise schedulers (like non-isotropic ones \cite{Curreli_2025_CVPR}) to capture more structural information.
While our experiments on real scans show that \methodname~is robust to real noise, it would likely fail in the presence of heavy clutter, e.g., furniture or large floor areas. Throughout our experiments, we also observed that methods trained on clean samples incur significant degradation when dealing with partial-view shapes, and \methodname~is no exception. This suggests that retraining is required to face heavy partiality, while more sophisticated, structured features might also be worth exploring \cite{human3d}.

\small{
\subsubsection{Acknowledgments}
This work is funded by the Deutsche Forschungsgemeinschaft - 409792180 (Emmy Noether Programme, project: Real Virtual Humans). 
G. Pons-Moll is a member of the Machine Learning Cluster of Excellence, EXC number 2064/1 – Project number 390727645. 
The authors thank the International Max Planck Research School for Intelligent Systems (IMPRS-IS) for supporting I.~A.~Petrov. 
This project has received funding from the European Union’s Horizon 2020 Research and Innovation Programme under the Marie Skłodowska-Curie Grant Agreement No. 101109330.
The project was made possible by funding from the Carl Zeiss Foundation.
This work was supported by the European Research Council (ERC) Advanced Grant SIMULACRON and by the GNI Project “AI4Twinning”. 
}

\bibliographystyle{splncs04}
\bibliography{029-main}

\input{029-supp}

\end{document}

%% file: 029-supp.tex
% This is samplepaper.tex, a sample chapter demonstrating the
% LLNCS macro package for Springer Computer Science proceedings;
% Version 2.24 (29-Jan-2024)

%
% Includes modifications for GCPR 2026, adapted from the GCPR 2024 template.
%
 
\newif\ifreview
% ANONYMOUS SUBMISSION FOR REVIEW
% \reviewtrue
% CAMERA READY SUBMISSION  
\reviewfalse

\ifreview
	\usepackage{lineno}
	\renewcommand\thelinenumber{\color[rgb]{0.2,0.5,0.8}\normalfont\sffamily\scriptsize\arabic{linenumber}\color[rgb]{0,0,0}}
	\renewcommand\makeLineNumber {\hss\thelinenumber\ \hspace{6mm} \rlap{\hskip\textwidth\ \hspace{6.5mm}\thelinenumber}} 
	\linenumbers
\fi

%%%%%%%%%%%%%%%%%%%%% Add submission id, track, and title. %%%%%%%%%%%%%%%%%%%%%

% TODO: Please insert your submission number here
\def\SubNumber{29}

% TODO: Please uncomment the track this paper will be submitted to, comment all other lines
\def\GCPRTrack{Main Track}
%\def\GCPRTrack{Special Track: Pattern recognition in the life and natural sciences}
%\def\GCPRTrack{Special Track: Photogrammetry and remote sensing}
%\def\GCPRTrack{Special Track: Computer vision systems and applications}
%\def\GCPRTrack{Young Researcher's Forum}
%\def\GCPRTrack{Fast Review Track}
%\def\GCPRTrack{Extended Abstract}

% TODO: Replace with your title
\title{Supplementary Materials of \\ Ordered Diffusion for 3D Human Registration}
% You can use \thanks for acknowledgment. Do not add any acknowledgment to the draft 
% version that is used for the review process.  
%\title{Title\thanks{XXX}}

\ifreview
	% ANONYMOUS SUBMISSION FOR REVIEW
	% DO NOT MODIFY these for the draft version that is used for the review process.
	\titlerunning{GCPR 2026 Submission \SubNumber{}. CONFIDENTIAL REVIEW COPY.}
	\authorrunning{GCPR 2026 Submission \SubNumber{}. CONFIDENTIAL REVIEW COPY.}
	\author{GCPR 2026 - \GCPRTrack{}}
	\institute{Paper ID \SubNumber}
\else
	% CAMERA READY SUBMISSION
	%\titlerunning{Abbreviated paper title}
	% If the paper title is too long for the running head, you can set
	% an abbreviated paper title here

    \author{Mattia Masiero\inst{1}\orcidlink{0009-0001-1327-0849} \and
    Ilya A. Petrov\inst{1,2}\orcidlink{0000-0002-8900-1071} \and \\
    Daniel Cremers\inst{3,4}\orcidlink{0000-0002-3079-7984} \and 
    Gerard Pons-Moll\inst{1,2,5}\orcidlink{0000-0001-5115-7794} \and
    Riccardo Marin\inst{3,4}\orcidlink{0000-0003-2392-4612}}
	
	\authorrunning{M. Masiero et al.}
	% First names are abbreviated in the running head.
	% If there are more than two authors, 'et al.' is used.
    \institute{
    $^{1}${\small University of T\"ubingen, Germany}\qquad
    $^{2}$ {\small T\"ubingen AI Center, Germany}\\
    $^{3}$ {\small Technical University of Munich, Germany}\\
    $^{4}$ {\small Munich Center for Machine Learning, Germany}\\
    $^{5}$ {\small Max Planck Institute for Informatics, Saarland Informatics Campus, Germany}\\
    \url{https://riccardomarin.github.io/odin/}}
\fi

\maketitle              % typeset the header of the contribution

\begin{abstract}
In this document, we provide implementation details of the diffusion model and feature extractor used in \methodname~. We also describe the post-prediction SMPL fitting procedures in detail. In addition, we present an experiment on the role of local conditioning during denoising, describe our data processing, and provide further experimental results.
\end{abstract}

\section{Architecture Details}

\subsection{Denoising Diffusion Model}

The architecture of the denoising model $D_{\phi}$, adapted from \cite{petrov2024tridi}, is based on a transformer. We use dedicated MLPs to embed the noised SMPL template vertex coordinates, local and global scan features, and a standard sinusoidal timestep embedding into a common 1024-dimensional space. This yields tokenized versions of the inputs. The tokens that encode vertex-specific information (template vertex tokens, local feature tokens, positional encoding) are summed for each vertex. The remaining conditioning sources relate to the sample as a whole (global feature token, timestep embedding) and are concatenated to the point representations. The resulting matrix is passed to a transformer encoder with 4 layers, 8 attention heads, and a feed-forward dimension of 512. The first 689 output tokens are decoded to 3D coordinates via an MLP. They represent the predicted noise added to the input coordinates, since the model is supervised using an L2 loss against the ground-truth noise. The noised template coordinates $\mathbf{x}_t$, local scan features $\mathbf{f}_{\text{local}}$, global scan features $\mathbf{f}_{\text{global}}$, and timestep embedding $\mathbf{t}_{\text{emb}}$ are independently mapped into a shared space of dimension $d=1024$ using MLPs:
\[
\hat{\mathbf{x}}_t^{i}=P_x(\mathbf{x}_t^{i}),
\qquad
\hat{\mathbf{f}}_{\text{local}}^{i}=P_l(\mathbf{f}_{\text{local}}^{i}),
\]

\[
\hat{\mathbf{f}}_{\text{global}}=P_g(\mathbf{f}_{\text{global}}),
\qquad
\hat{\mathbf{t}}_{\text{emb}}=P_t(\mathbf{t}_{\text{emb}}).
\]
Vertex tokens are constructed as:

\[
\mathbf{v}^i
=
\hat{\mathbf{x}}_t^{i}
+
\hat{\mathbf{f}}_{\text{local}}^{i}
+
\mathrm{PE}(i),
\qquad
i=1,\ldots,689.
\]
The transformer input is

\[
\mathbf{e}
=
\left[
\mathbf{v}^{1},
\dots,
\mathbf{v}^{689},
\hat{\mathbf{f}}_{\text{global}},
\hat{\mathbf{t}}_{\text{emb}}
\right]^\top
\in
\mathbb{R}^{691\times1024}.
\]
After layer normalization, the sequence is processed by the transformer encoder:

\[
\mathbf{y}
=
\mathrm{Transformer}
\bigl(
\mathrm{LN}(\mathbf{e})
\bigr)
\in
\mathbb{R}^{691\times1024}.
\]
The first $689$ output tokens are decoded into 3D coordinates and interpreted as the predicted diffusion noise:

\[
\hat{\boldsymbol{\epsilon}}
=
P_\epsilon\!\left(
\mathbf{y}_{1:689}
\right)
\in
\mathbb{R}^{689\times3}.
\]

\begin{table}[t]
\centering
\caption{Architecture of the denoising transformer.}
\begin{tabular}{lll}
\hline
Symbol & Module & Output Shape \\
\hline

$P_x$ &
Linear(3,1024) $\rightarrow$ SiLU $\rightarrow$ Linear(1024,1024)
&
$(B,689,1024)$
\\

$P_l$ &
Linear(128,1024) $\rightarrow$ SiLU $\rightarrow$ Linear(1024,1024)
&
$(B,689,1024)$
\\

$P_g$ &
Linear(256,1024) $\rightarrow$ SiLU $\rightarrow$ Linear(1024,1024)
&
$(B,1024)$
\\

$P_t$ &
Linear(1024,1024) $\rightarrow$ SiLU $\rightarrow$ Linear(1024,1024)
&
$(B,1024)$
\\

$\mathrm{PE}$ &
Sinusoidal positional encoding
&
$(689,1024)$
\\

$\mathrm{LN}$ &
LayerNorm(1024)
&
$(B,691,1024)$
\\

Transformer &
4 layers, 8 heads, FFN=512, dropout=0.1
&
$(B,691,1024)$
\\

$P_\epsilon$ &
Linear(1024,3) $\rightarrow$ SiLU $\rightarrow$ Linear(3,3)
&
$(B,689,3)$
\\

\hline
\end{tabular}
\label{tab:denoiser_architecture}
\end{table}

\subsection{Feature Extractor}

We use a PointNet++ \cite{qi2017pointnet++} architecture to extract features from the input point cloud. We adapt the implementation from \cite{Pytorch_Pointnet_Pointnet2}.
The input scan point cloud $\mathbf{S}\in\mathbb{R}^{4096\times3}$ is internally transposed to $\mathbf{S}^{\top}\in\mathbb{R}^{3\times4096}$ to conform to the implementation. The network then processes the scan through four hierarchical Set Abstraction (SA) layers, progressively reducing the resolution while increasing the feature dimensionality. The feature representation at the network's bottleneck is the most general and compact. Therefore, it is the obvious choice for extracting a global scan feature.

Since the denoising transformer demands a single global conditioning vector per scan, we modify the SA4 layer to meet this requirement. In the standard formulation, SA4 outputs features of shape $(c,f)$, where $c$ is the number of centroids and $f=\sum_i f_i$ is the combined output dimensionality of the scale-specific MLPs. Instead of aggregating multiple centroids, e.g., via a pooling operation, we set $c=1$ and choose the MLP output dimensions such that $f=256$. This yields a feature of dimensionality $(1,256)$, which can be unambiguously reshaped to a global feature vector $\mathbf{f}_{\mathrm{global}}\in\mathbb{R}^{256}$.

After the bottleneck, Feature Propagation (FP) layers upsample the features back to the original scan resolution. The output of the final feature propagation layer, FP1, yields the local scan features $\mathbf{f}_{\mathrm{local}}\in\mathbb{R}^{4096\times128}$.

The final part of the architecture is the segmentation head. In our case, however, we do not use the network for standalone segmentation; instead, the feature extractor is trained end-to-end with the diffusion model.

\begin{table}[t]
\centering
\caption{Architecture of the PointNet++ scan feature extractor.}
\begin{tabular}{lll}
\hline
Module & Configuration & Output Shape \\
\hline

SA$_1$ &
$n=1024$
&
$(B,96,1024)$
\\
&
radii $(0.05,0.1)$
&
\\
&
samples $(16,32)$
&
\\
&
MLPs $[16,16,32]$, $[32,32,64]$
&
\\
\hline

SA$_2$ &
$n=256$
&
$(B,256,256)$
\\
&
radii $(0.1,0.2)$
&
\\
&
samples $(16,32)$
&
\\
&
MLPs $[64,64,128]$, $[64,96,128]$
&
\\
\hline

SA$_3$ &
$n=64$
&
$(B,512,64)$
\\
&
radii $(0.2,0.4)$
&
\\
&
samples $(16,32)$
&
\\
&
MLPs $[128,196,256]$, $[128,196,256]$
&
\\
\hline

SA$_4$ $ \rightarrow \mathbf{f}_{\mathrm{global}}$ &
$n=1$
&
$(B,256,1)$
\\
&
radii $(0.4,0.8)$
&
\\
&
samples $(16,32)$
&
\\
&
MLPs $[128,128]$, $[128,128]$
&

%&
%$\mathbf{f}_{\mathrm{global}}\in\mathbb{R}^{256}$
\\
\hline

FP$_4$ &
input channels $768$
&
$(B,256,64)$
\\
&
MLP $[256,256]$
&
\\
\hline

FP$_3$ &
input channels $512$
&
$(B,256,256)$
\\
&
MLP $[256,256]$
&
\\
\hline

FP$_2$ &
input channels $352$
&
$(B,128,1024)$
\\
&
MLP $[256,128]$
&
\\
\hline

FP$_1$ $\rightarrow \mathbf{f}_{\mathrm{local}}$ &
input channels $128$
&
$(B,128,4096)$
\\
&
MLP $[128,128,128]$
&
\\
\hline

Segmentation head &
Conv(128,128)
&
$(B,24,4096)$
\\
&
BN $\rightarrow$ ReLU
&
\\
&
Dropout(0.5)
&
\\
&
Conv(128,24)
&
\\

\hline
\end{tabular}
\label{tab:pointnet_scan_feature_extractor}
\end{table}

\section{SMPL Fitting}
In this section, we present additional details about the SMPL fitting stages of our method. We use a gender-neutral SMPL body model, as gender information is unavailable at inference time.

\subsection{Fitting SMPL to Diffusion Output}
We use smplfitter \cite{sarandi24nlf} to recover the SMPL instance closest to the ordered point cloud produced by the diffusion model. This requires the predicted point cloud to be in semantic correspondence with SMPL, a property that is preserved by our diffusion formulation. Essentially, this step consists of inverting the parameter-to-mesh mapping that is SMPL. In our setting, we only have a subset of SMPL vertices (689 out of 6890) and no joint locations. While smplfitter supports this scenario, it would normally require a custom joint regressor matrix to infer joint locations from the reduced vertex set. We address this by selecting the 689 template vertices so that they contain all the vertices involved in the original SMPL joint regression. Thus, the standard SMPL joint regressor can still be applied directly, eliminating the need to estimate a new regressor for the reduced template.

We perform two iterations with smplfitter and use a shape regularization weight of $10^{-4}$.

\subsection{Fitting SMPL to the Scan}
\label{subsec:fitting_smpl_to_scan}

Finally, we refine the SMPL parameters to directly match the input scan. This optimization procedure is adapted from \cite{marin_nicp_2025}. Thanks to the preceding SMPL optimization stage, we begin with a SMPL instance that is already close to the registered template and, assuming the template is a good fit, also close to the scan. This high-quality initialization is essential, as the optimization in this stage relies on the chamfer distance, which is prone to local minima and degenerate solutions. Specifically, we use the single-directional chamfer distance from the scan to the optimization SMPL model. We use an initial learning rate of 0.02, which decays linearly to zero over 1000 iterations. The loss consists of the chamfer distance from the scan to the SMPL model vertices, an L2 shape regularization term weighted with 0.2, as well as a pose regularization term implemented with VPoser \cite{pavlakos2019expressive}, weighted with 0.001. In the following subsection, we discuss VPoser regularization in greater detail.

\subsection{Pose Regularization}
\label{subsec:vposer}
Variational Human Body Pose Prior (VPoser), \cite{pavlakos2019expressive} is a body pose prior for the SMPL model. It features a Variational Autoencoder (VAE) trained for pose reconstruction on a diverse dataset of human poses. As a result of its training objective, it imposes a standard Gaussian distribution on its latent space. We can leverage this property to quantify the likelihood of a given pose. To do so, we first encode the pose using the VPoser encoder, which outputs a latent distribution parameterized by a mean and a variance. We consider the mean because it represents the most likely latent code for the input pose. Since the prior distribution is standard normal, its mode lies at the origin. Therefore, the L2 norm of the latent mean serves as a proxy for pose likelihood: larger values correspond to less likely poses. During SMPL optimization, we leverage this to discourage unlikely poses.

\section{Data Processing}
\label{appendix:dataset_preprocessing}

Here we provide additional details on the data processing. For consistency and computational efficiency, all scans are represented by 4096 points. We scale the input point clouds by a factor of 4.5 relative to their original SMPL scale to match the signal-to-noise ratio used in the diffusion implementation of \cite{melas2023pc2}, which serves as the basis for our implementation.

\subsection{AMASS}
\label{sec:amass}
For AMASS \cite{mahmood2019amass} samples, we uniformly sample a point cloud of 4096 points from the surface of the SMPL mesh. This serves as the input scan. Gaussian noise with standard deviation $0.0025$ is added to the virtual scan points to simulate sensor noise. The corresponding ground-truth registration is represented by the SMPL template vertices. Both the scan and the template vertices are centered using the centroid of the selected 689 template vertices used by the diffusion model. This ensures that the diffusion vertices are mean-zero, as is the Gaussian noise in the diffusion framework, and that the alignment between the ground truth and the scan is preserved. Optionally, we can apply two types of augmentation to the virtual scan point clouds, as detailed below. When we refer to augmentations on the train set, a probabilistic ratio of 0.75:0.25 of unaffected to augmented samples is used. For evaluation on augmented test sets, every sample is augmented.

\paragraph{Missing Limbs.}
When a scan is selected for missing-limb augmentation, exactly one of the four outer limbs (outer left arm, right arm, left leg, or right leg) is chosen uniformly at random, and all scan points belonging to the corresponding body part are removed. We use the canonical SMPL segmentation into 24 body parts to identify the affected points. To preserve a constant scan size of 4096 points, the remaining points are resampled with replacement until the original point count is restored.

\paragraph{Partial Views.}
For this augmentation, a virtual camera is placed at a random azimuth angle around the subject. Surface points are densely (5*4096) sampled from the underlying mesh, and only points whose normals face the camera are retained, approximating the visible surface from that viewpoint. A set of 4096 points is then sampled from the visible region; if fewer than 4096 visible points are available, the remainder is obtained by resampling with replacement. 

\subsection{DFAUST}
For the DFAUST \cite{bogo2017dynamic} dataset we adopt the train/test split used in \cite{feng2023generalizing} as well as \cite{chen2021snarf}. Moreover, we apply the same frame subsampling as in \cite{feng2023generalizing}: first, we restrict the selection to the middle 50\% of frames in each sequence, since each sequence begins and ends with the subject in a standardized A-pose. Second, we keep only every fifth frame because the high frame rate introduces temporal redundancy. Together, this yields about 3800 data points for training and 313 for testing. We use the provided scan meshes together with their corresponding ground-truth registrations. To remove outlier noise, the scans are filtered by removing vertices whose distance to the registration exceeds $0.1$ m. The filtered scan meshes are then converted into point clouds by first sampling 50,000 points uniformly from their surfaces and then applying farthest-point sampling to obtain 4096 points. The data is centered as described in Sec.\ref{sec:amass}.

\subsection{FAUST}
FAUST \cite{bogo_faust_2014} contains individual pairs of scans and ground-truth SMPL fits. Since only the train split provides ground-truth fits, we restrict our evaluation to this subset. We filter the scan meshes by removing vertices whose distance to the corresponding registration exceeds $0.1$ m. The filtered scan meshes are then converted into point clouds by uniformly sampling 4096 points from their surfaces. The data is centered as described in the AMASS section.

\section{Role of Local Conditioning}

\begin{figure}[t]
    \centering
    \includegraphics[width=0.9\linewidth]{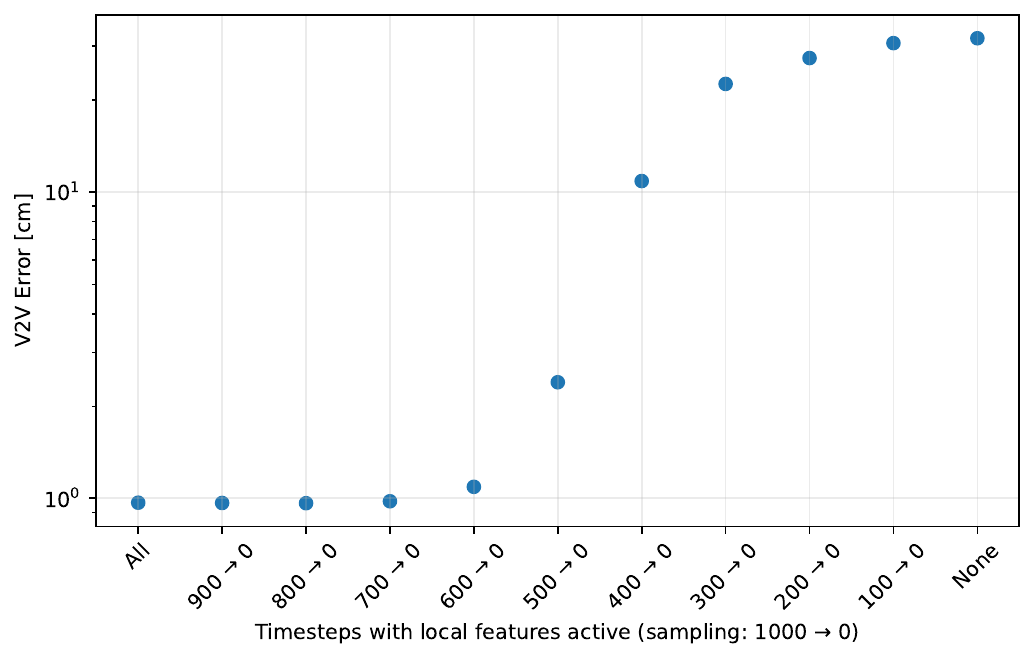}
    \caption{Effect of disabling local conditioning up to different cutoff timesteps.}
    \label{fig:local_feature_cutoff}
\end{figure}

\paragraph{Experimental Design.} In the ablation experiment, we have seen that the local scan features are crucial for accurate reconstruction. However, at high-noise timesteps ($t\approx1000$), nearest-neighbor feature assignments are essentially arbitrary. This raises the question of how the denoiser uses the local conditioning throughout the diffusion process. Hence, we examine the effect of deactivating local conditioning at test time, from the start of sampling ($t=1000$) until select cutoff timesteps $\tau \in \{1000 , 900 ..., 0\}$. We employ our regular model, which has been trained with nearest-neighbor local feature assignment at every timestep. Consequently, $\tau=1000$ corresponds to the unmodified model, whereas $\tau=0$ disables local conditioning throughout all of sampling. During the affected interval, the local conditioning mechanism is deactivated by replacing the assigned local features with a fixed dataset-level mean. After the affected interval, the original nearest-neighbor assignments are resumed. Performance is measured before SMPL fitting using the V2V error between the diffusion sample and the subsampled ground-truth point cloud.

\paragraph{Results.}
We observe that deactivating local conditioning during the early diffusion timesteps ($t=1000 \rightarrow 700$) has a negligible ($<2\%$) effect on reconstruction quality. This suggests that the denoiser places little reliance on local conditioning while it provides a largely random signal. For $\tau=600$, the V2V error increases by approximately $13\%$. For the subsequent values of $\tau$, the V2V error degrades severely, plateauing at over $30\times$ the original error. At these timesteps, replacing local conditioning with an uninformative placeholder substantially degrades the reconstruction, indicating that the denoiser increasingly relies on local conditioning as denoising progresses. These observations suggest that the contribution of local conditioning increases over the course of denoising. As the generated template becomes better aligned with the target scan, nearest-neighbor assignments become more reliable, making local conditioning increasingly informative.

\section{Further Qualitative Results}
In the following, we report further qualitative comparisons on FAUST~\cite{bogo_faust_2014}, more on missing-limbs completion on DFAUST~\cite{bogo2017dynamic}, results of ODin on highly out-of-distribution poses from SHREC19~\cite{melzi2019shrec} (which include shapes from TOSCA \cite{bronstein2008numerical}, SCAPE~\cite{anguelov2005scape}, and test scans from FAUST).

\begin{figure}
\begin{overpic}[width=\linewidth, trim=0 20 0 0, clip]{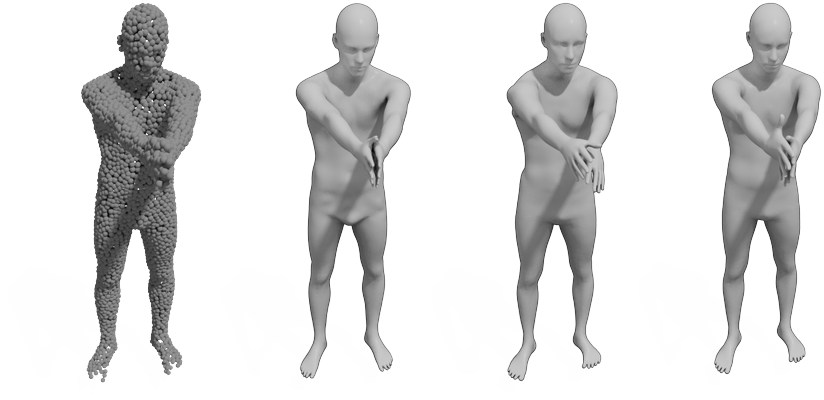}
    \put(15, 47){Input}
    \put(41, 47){GT}
    \put(66, 47){NICP}
    \put(90, 47){\textbf{ODin}}
\end{overpic}
\caption{Comparison on FAUST}
\end{figure}

\begin{figure}
\begin{overpic}[width=\linewidth, trim=0 20 0 0, clip]{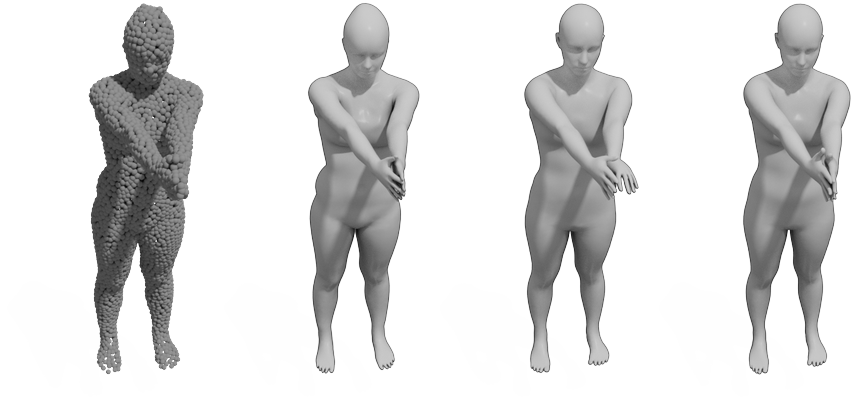}
    \put(15, 45){Input}
    \put(41, 45){GT}
    \put(66, 45){NICP}
    \put(90, 45){\textbf{ODin}}
\end{overpic}
\caption{Comparison on FAUST}
\end{figure}

\begin{figure}
\begin{overpic}[width=\linewidth, trim=0 20 0 0, clip]{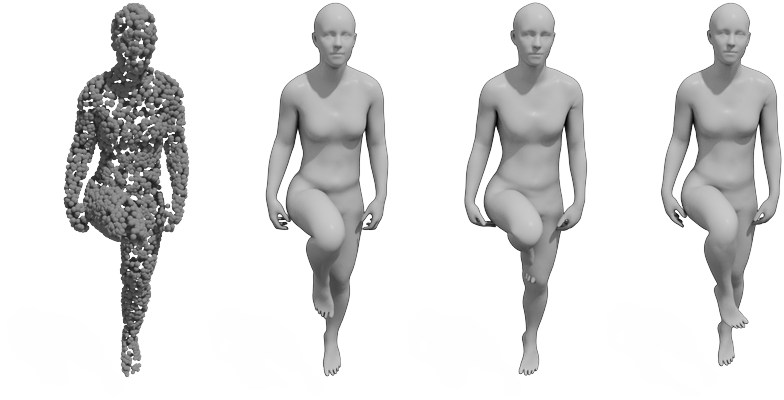}
    \put(15, 50){Input}
    \put(41, 50){GT}
    \put(66, 50){NICP}
    \put(90, 50){\textbf{ODin}}
\end{overpic}
\caption{Comparison on missing limbs}
\end{figure}

\begin{figure}
\begin{overpic}[width=\linewidth, trim=0 20 0 0, clip]{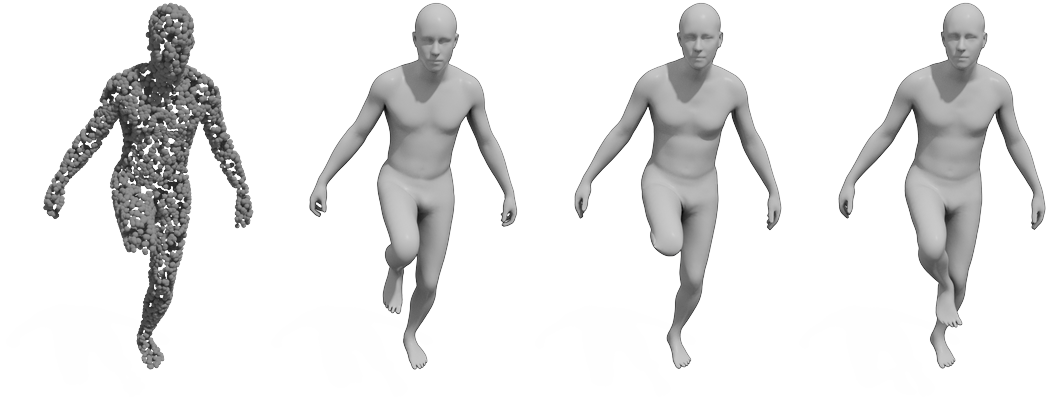}
    \put(15, 37){Input}
    \put(41, 37){GT}
    \put(66, 37){NICP}
    \put(90, 37){\textbf{ODin}}
\end{overpic}
\caption{Comparison on missing limbs}
\end{figure}

\begin{figure}
\begin{overpic}[width=\linewidth, trim=0 20 0 0, clip]{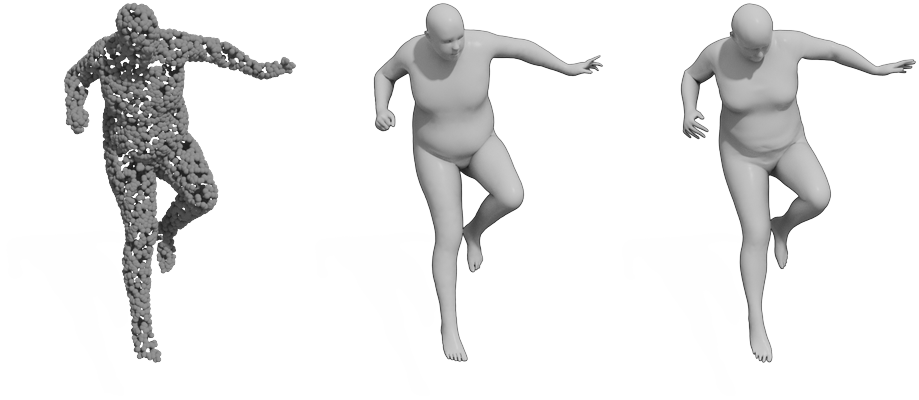}
    \put(15, 43){Input}
    \put(45, 43){GT}
    \put(75, 43){\textbf{ODin}}
\end{overpic}
\caption{Results on SHREC19}
\end{figure}

\begin{figure}
\begin{overpic}[width=\linewidth, trim=0 20 0 0, clip]{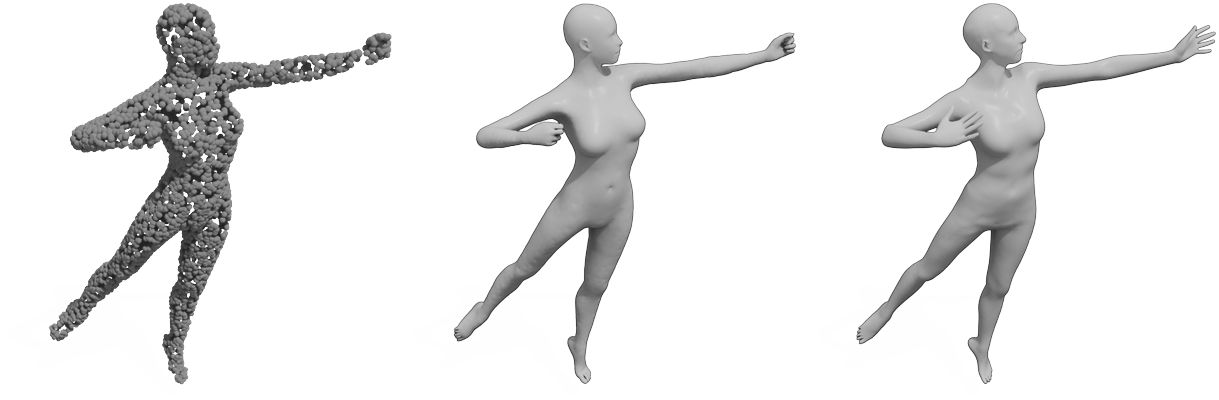}
    \put(15, 35){Input}
    \put(45, 35){GT}
    \put(75, 35){\textbf{ODin}}
\end{overpic}
\caption{Results on SHREC19}
\end{figure}

\begin{figure}
\begin{overpic}[width=\linewidth, trim=0 20 0 0, clip]{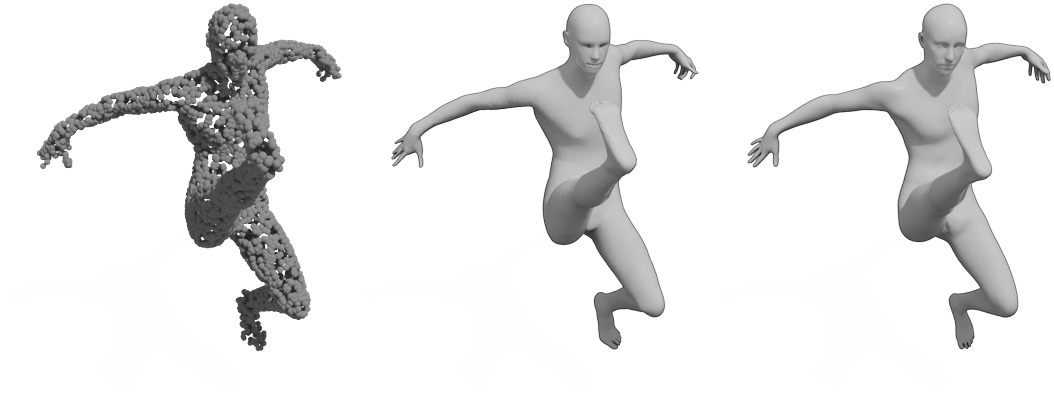}
    \put(16, 41){Input}
    \put(49, 41){GT}
    \put(81, 41){\textbf{ODin}}
\end{overpic}
\caption{Results on SHREC19}
\end{figure}

\begin{figure}
\begin{overpic}[width=\linewidth, trim=0 20 0 0, clip]{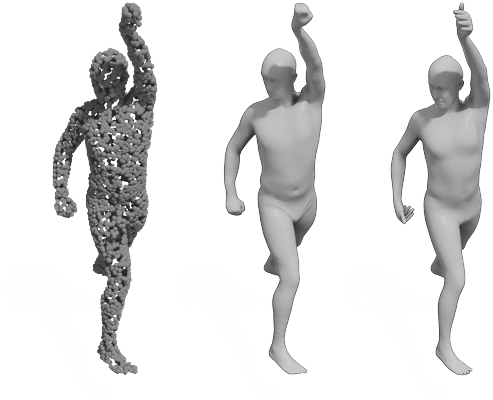}
    \put(15, 70){Input}
    \put(49, 70){GT}
    \put(80, 70){\textbf{ODin}}
\end{overpic}
\caption{Results on SHREC19}
\end{figure}

\begin{figure}
\begin{overpic}[width=\linewidth, trim=0 20 0 0, clip]{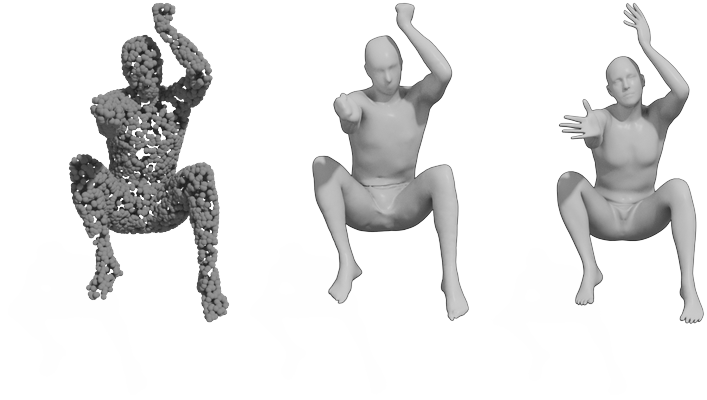}
    \put(15, 50){Input}
    \put(48, 50){GT}
    \put(78, 50){\textbf{ODin}}
\end{overpic}
\caption{Results on SHREC19}
\end{figure}

\begin{figure}
\begin{overpic}[width=\linewidth, trim=0 20 0 0, clip]{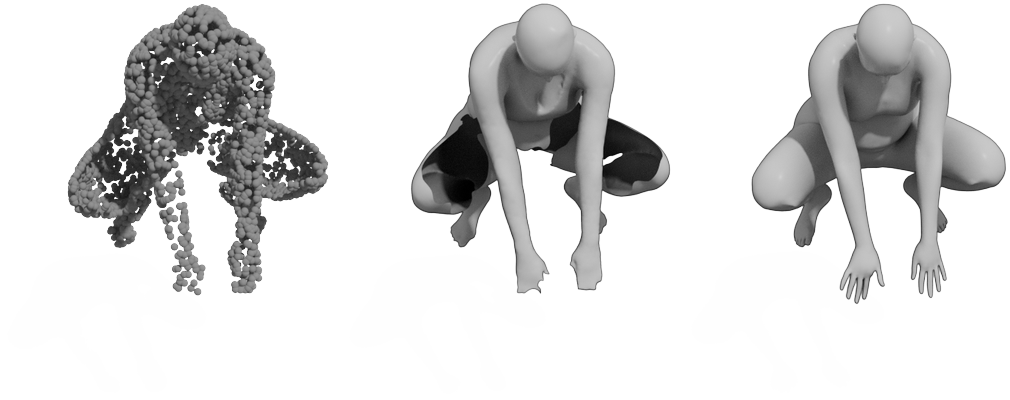}
    \put(18, 40){Input}
    \put(48, 40){GT}
    \put(78, 40){\textbf{ODin}}
\end{overpic}
\caption{Results on SHREC19}
\end{figure}

\clearpage
\newpage